\documentclass[conference]{IEEEtran}

\usepackage[utf8]{inputenc}
\usepackage{cite}
\usepackage[linesnumbered,vlined,ruled,boxed,commentsnumbered]{algorithm2e}
\usepackage{algpseudocode}
\usepackage{bm} 

\usepackage{amsmath,amssymb,amsfonts}
\usepackage{commath}
\usepackage{float}

\usepackage[flushleft]{threeparttable}
\usepackage{cleveref}

\usepackage{graphics,graphicx}
\usepackage{subfigure}

\usepackage[acronym,smallcaps]{glossaries}

\newcommand{\fref}[1]{Fig. \ref{#1}}
\newcommand{\tref}[1]{Table \ref{#1}}

\newcommand{\sref}[1]{Section \ref{#1}}
\newcommand{\eref}[1]{Eq. \ref{#1}}
\newcommand{\aref}[1]{Algorithm \ref{#1}}

\usepackage{xcolor} %

\newacronym{larfdssom}{LARFDSSOM}{\textit{Local
Adaptive Receptive Field Dimension Selective Self-organizing Map}}
\newacronym{soms}{SOMs}{\textit{Self-Organizing Maps}}
\newacronym{som}{SOM}{\textit{Self-Organizing Map}}
\newacronym{larfsom}{LARFSOM}{\textit{Local Adaptive Receptive Field Self-Organizing Map}}
\newacronym{dssom}{DSSOM}{\textit{Dimension Selective Self-Organizing Map}}
\newacronym{sssom}{SS-SOM}{\textit{Semi-Supervised Self-Organizing Map}}
\newacronym{ssl}{SSL}{\textit{Semi-Supervised Learning}}
\newacronym{lvq}{LVQ}{\textit{Learning Vector Quantization}}
\newacronym{s3vm}{$S^3VM$}{\textit{Semi-supervised Support Vector Machines}}
\newacronym{svm}{SVM}{\textit{Support Vector Machines}}
\newacronym{hmrf}{HMRFs}{\textit{Hidden Markov Random Fields}}
\newacronym{em}{EM}{\textit{Expectation Maximization}}
\newacronym{lp}{LP}{\textit{Label Propagation}}
\newacronym{ijcnn}{IJCNN}{\textit{International Joint Conference on Neural Networks}}
\newacronym{mlp}{MLP}{\textit{Multilayer Perceptron}}
\newacronym{grlvq}{GRLVQ}{\textit{Generalized Relevance Learning Vector Quantization}}
\newacronym{ls}{LS}{\textit{Label Spreading}}
\newacronym{ce}{CE}{\textit{Clustering Error}}
\newacronym{lhs}{LHS}{\textit{Latin Hypercube Sampling}}
\newacronym{csom}{CSOM}{\textit{Convolutional Self-Organizing Map}}
\newacronym{dsom}{DSOM}{\textit{Deep Self-Organizing Map}}
\newacronym{doc}{DOC}{\textit{Densitive-based Optimal projective Clustering}}
\newacronym{proclus}{PROCLUS}{\textit{PROjected CLUStering algorithm}}

\newacronym{propm}{ALTSS-SOM}{\textit{Adaptive Local Thresholds Semi-Supervised Self-Organizing Map}}

\begin{document}

\title{A Semi-Supervised Self-Organizing Map with Adaptive Local Thresholds}
\author{
  \IEEEauthorblockN{Pedro H. M. Braga, \textit{Member}, \textit{IEEE}, and Hansenclever F. Bassani, \textit{Member}, \textit{IEEE}}
  \IEEEauthorblockA{Center of Informatics - CIn, Universidade Federal de Pernambuco, Recife, PE, Brazil, 50.740-560\\
  Email: \{phmb4, hfb\}@cin.ufpe.br}
}

\maketitle

\begin{abstract}
In the recent years, there is a growing interest in semi-supervised learning, since, in many learning tasks, there is a plentiful supply of unlabeled data, but insufficient labeled ones. Hence, Semi-Supervised learning models can benefit from both types of data to improve the obtained performance. Also, it is important to develop methods that are easy to parameterize in a way that is robust to the different characteristics of the data at hand. This article presents a new method based on Self-Organizing Map (SOM) for clustering and classification, called \gls{propm}. It can dynamically switch between two forms of learning at training time, according to the availability of labels, as in previous models, and can automatically adjust itself to the local variance observed in each data cluster. The results show that the \gls{propm} surpass the performance of other semi-supervised methods in terms of classification, and other pure clustering methods when there are no labels available, being also less sensitive than previous methods to the parameters values. 
\end{abstract}

\IEEEpeerreviewmaketitle

\section{Introduction}
\label{sec:intro}

Over the last few years, the use of machine-learning technology has driven many aspects of modern society. Recent research on Artificial Neural Networks with supervised learning has shown great advances. It is the most common form of machine learning \cite{lecun2015deep}. It is not unusual to see on the news several practical applications, in diverse areas \cite{lecun2015deep,sssom}. A key to the success of supervised learning, especially, deep supervised learning, is the availability of sufficiently large amounts of labeled training data. Unfortunately, creating such properly labeled data with enough examples for each class is not easy. As a result, the use of supervised learning methods became impractical in many applications such as in the medical field, where it is extremely difficult and expensive to obtain balanced labeled data.

On the other hand, due to the advances in technology that have produced datasets of increasing size, in terms of the number of samples and features, unlabeled data usually can be easily obtained. Therefore, it is of great importance to put forward methods that can combine both types of data in order to benefit from the information they can provide, each of them in their way \cite{chapelle2009semi}. An approach typically applied in such scenario is \gls{ssl}. It is a halfway between supervised and unsupervised learning and can be used to both clustering and classification tasks \cite{chapelle2009semi,schwenker2014pattern}.

We point out that prototype-based methods have been successfully applied for both tasks. Methods based on Self-Organizing Maps \cite{kohonen1990, bassani2015larfdssom,sssom} and K-Means \cite{zhu2002-label-propagation} can be highlighted as examples, as well as deep learning techniques \cite{csom,chen2018semi,rasmus2015semi}. The \gls{som} is an unsupervised learning method, frequently applied for clustering, while \gls{lvq} \cite{kohonen1990}, its supervised counterpart that shares many similarities, is normally used for classification. They both were proposed by Kohonen, and since then, various modifications have been introduced, including semi-supervised versions \cite{sssom}, to deal with more challenging problems.

Recent \gls{som}-based methods employ a threshold defining the minimum level of activation for an input pattern to be considered associated with a cluster prototype. This threshold level is a parameter of the model which is shared by all prototypes \cite{bassani2015larfdssom, sssom}, thus, the regions that a prototype can represent are not learned at all, or they are normally estimated using supervised approaches, as in \cite{fischer2016optimal}.

In this context, the main idea of this paper is to introduce the concept of local adaptive thresholds through the use of the local variances observed by the prototypes for each dimension. Such variances are calculated using a bias-corrected moving average with an exponentially weighted decay rate \cite{adam}. This concept was derived from the idea of rejection options, early introduced by Chow \cite{chow}. It is related to the conditions of taking a classification or recognition decision for a particular point or a data region. In the case of \gls{som}-based methods, these decisions are associated with the nodes in the map (i.e., when they must accept an input pattern to be part of its representation pool). Such rejection options define the first step towards an adaptation of the model complexity tailored to data regions with a high degree of uncertainty \cite{fischer2014rejection}. So far, most models that use rejection options deals with just a single threshold, as well as most of them can handle only with binary classification \cite{fischer2016optimal}.

In this article, we propose a new model called \gls{propm}, which is an extension of \gls{sssom}, created by introducing important modifications to incorporate the ability to estimate local rejection options as a function of both local variance and relevance of the input dimensions for each node in the map. To evaluate \gls{propm}, we compared it with other semi-supervised approaches that do not use adaptive reject options. We also compare \gls{propm} with its predecessor that used a parameter to define the threshold region to make pattern rejection decisions. Also, once we introduce an entirely new learning procedure, it becomes necessary to compare \gls{propm} not only regarding the classification rate but also considering the clustering rate. It is done by taking into consideration the methods that provided the ideas for the development of the proposed method as well as other conventional methods in the literature. Finally, as our parameter sensitivity analysis shows, the sensibility of the model to the parameters was significantly decreased in comparison with the previous version.

The rest of this article is organizing as follows: \sref{sec:background} presents a short review of the background related to the areas where this paper is inserted. \sref{sec:related-work} introduces related work in the literature. \sref{sec:prop-method} describes in detail the proposed method. \sref{sec:experiments} presents the experimental setup, methodology, the obtained results, and comparisons. Finally, \sref{sec:conclusions} discusses the obtained results and draws the conclusion of this paper, as well as indicates future directions.

\section{Background}
\label{sec:background}

High-dimensional data poses different challenges for clustering tasks. In particular, similarity measures used in traditional clustering techniques may become meaningless due to the curse of dimensionality \cite{koppen2000curse}. In this context, subspace clustering and projected clustering appear as common choices. They aim to determine clusters in subspaces of the input dimensions. This task involves not only the clustering itself but also identifying relevant subsets in the input dimensions for each cluster \cite{kriegel2009clustering}. One way to achieve this is by applying local relevances to the input dimensions.

Moreover, as in \cite{sssom}, this paper introduces a model that is able not only to cluster but also classify samples. In this context, we aim to introduce the concept of reject options. According to Chow \cite{chow}, uncertainties and noise inherent in any pattern recognition task result in a certain amount of unavoidable errors. Uncertainty normally has two reasons: points being outliers or located in ambiguous regions \cite{vailaya2000reject}. The option to reject is introduced to avoid an excessive misrecognition rate by converting errors into rejection.

We derive this idea to consider local reject options based on the variances estimated for each node in the map (discussed in more details further in the \sref{sec:prop-method}). The main idea is to give nodes the ability to reject an input pattern \textbf{x} if it is outside a region in the space defined from the estimated variances in each dimension around the centroid of each node. This will result in an adaptive local thresholding approach, similar to the one found in \cite{adaptive-local-thresholding, singh2012new}. However, this variance based approach provides a threshold adjusted for each dimension of each node in the map during the semi-supervised learning process.

Finally, it is important mentioning that \gls{ssl} perfectly fits for all of the referred problems and techniques. Because of that, considering the growing interest in semi-supervised learning in the past years, such combined approaches may come to arise more often. It is also worth pointing out that such interest for \gls{ssl} is growing in the machine-learning \cite{zhu2002-label-propagation,label-spreading,sssom} alongside in the deep learning context \cite{deep-ssl,dsom}.

\section{Related Work}
\label{sec:related-work}
This section briefly summarizes related works in diverse contexts. First, in the purely semi-supervised machine-learning context, we highlight the K-means based methods. K-means is one of the most popular and simple clustering algorithms that was proposed over 50 years ago, but it is still widely used in diverse applications \cite{basu2002semi,jain2010data}.

Continuing talking about prototype-based methods, the \gls{sssom} \cite{sssom} appears as the inspiration to the model presented in this paper. \gls{sssom} is a semi-supervised method that essentially inherits characteristics from its predecessor, \gls{larfdssom} \cite{bassani2015larfdssom}, but also introduces elements of supervision (i.e. supervised learning) to create a hybrid environment where \gls{ssl} could be applied. The \gls{larfdssom} \cite{bassani2015larfdssom} uses a time-varying structure, a neighborhood defined by connecting nodes that have similar subspaces of the input dimensions, and a local receptive field that is adjusted for each node as a function of its local variance. Both \gls{larfdssom} and \gls{sssom} were developed to deal with high-dimensional data, considering different learning contexts. The latter carries the ability to perform not only clustering tasks but also classification. Note that they works exactly as the same when there is no labeled sample available.

Label Propagation methods \cite{zhu2002-label-propagation} are another approach for \gls{ssl}. Essentially, they operate on proximity graphs or connected structures to spread and propagate class information to nearby nodes according to a similarity matrix. It is based on the assumption that nearby entities should belong to the same class, in contrast to far away entities. Label Spreading \cite{label-spreading} methods are very similar. The difference consists of modifications to the similarity matrix. The first uses the raw similarity matrix constructed from the data with no changes, whereas the latter minimizes a loss function that has regularization properties allowing it to be often better regarding robustness to noise.

Furthermore, in the literature, some state of the art strategies for rejection option can be listed \cite{fischer2014rejection}. On considering both local and global rejection, with the latter being the most common form, they can be divided into three distinct categories \cite{fischer2016optimal}: 1) probabilistic approaches; 2) turning deterministic measures into probabilities, and 3) deterministic approaches. Further, some adaptive local thresholding techniques are also found, like as in \cite{adaptive-local-thresholding}.

Moreover, it is possible to see in \cite{kriegel2009clustering} a review of methods that work well for problems of clustering in high-dimensional data. On considering this, and also the detailed comparison of subspace and projected clustering methods perform by \cite{bassani2015larfdssom}, we highlight \gls{proclus} \cite{proclus}, \gls{doc} \cite{doc} and the \gls{larfdssom} models due their good performances. \gls{doc} \cite{doc} is a cell-based method that searches for sets of grid cells containing more than a certain number of objects by using a Monte Carlo based approach that computes, with high probability, a good approximation of an optimal projective cluster. \gls{proclus} \cite{proclus} is a clustering-oriented algorithm that aims to find clusters in small projected subspaces by optimizing an objective function of the entire set of clusters, such as the number of clusters, average dimensionality, or other statistical properties.

\section{Proposed Method}
\label{sec:prop-method}

\gls{propm}\footnote{Available at: https://github.com/phbraga/alt-sssom} is a \gls{som} with Adaptive Local Thresholds \cite{adaptive-local-thresholding} based on \gls{sssom}. Hence, being based on \gls{sssom}, \gls{propm} can also learn in a supervised or unsupervised way depending on the availability of labels, and maintains the general characteristics of its predecessors. However, it introduces new supervised and unsupervised behaviors to allow a better usage, and consequently a better understanding of the data statistics. By doing this, \gls{propm} aims at overcoming the problems presented by \gls{sssom}, such as the high sensitivity to the parameters, and the low sample efficiency. Additionally, with the proper changes, \gls{propm} targets achieving better results for both classification and clustering tasks.

Therefore, the parameterized activation threshold ($a_t$) used in both previous methods is replaced by an adaptive thresholding technique that takes into account the local variance to provide the model the ability to learn the receptive field of each node. The objective is to estimate optimal local regions in the space with respect to the distribuition of the input patterns $\boldsymbol{x}$ for each node in the map. To do so, inspired by the Adam algorithm, a method for efficient stochastic optimization that only requires first-order gradients with little memory requirement \cite{adam}, \gls{propm} updates exponential moving averages of the distances between each input pattern $\boldsymbol{x}$ and the centroid of the nodes for each dimension (the vector $\boldsymbol{\delta}_{j}$ in the algorithms). In \gls{sssom} and \gls{larfdssom}, this estimate was done by using not only $\beta$ but also the learning rate $e$. However, \gls{propm} modified this approach to use solely the parameter $\beta \in \left[0,1\right)$ for controling the exponential decay rate of the moving averages.

The moving averages themselves are estimates of the first moment (the mean) of the distances between the input patterns and the centroids of the nodes. Because of that, such means can be used as estimates of the uncentered variance of the nodes in each dimension. However, these moving averages are initialized as vectors of zeros, leading to moment estimates that are biased towards zero, especially during the initial steps, and when the decay rate is small (close to 1) \cite{adam}. Still, according to \cite{adam}, this initialization bias can be counteracted, resulting in the bias-corrected estimate $\boldsymbol{\widehat{\delta_{j}}}$.

The moving averages themselves are estimates of the first moment (the mean) of the distances between the input patterns and the centroids of the nodes. Because of that, such means can be used as estimates of the uncentered variance of the nodes in each dimension. However, these moving averages are initialized as vectors of zeros, leading to moment estimates that are biased towards zero, especially during the initial steps, and when the decay rate is small (close to 1) \cite{adam}. Still, according to \cite{adam}, this initialization bias can be counteracted, resulting in a bias-corrected estimate $\hat{\boldsymbol{\delta}}_{j}$. During the learning process, this bias-corrected estimate $\hat{\boldsymbol{\delta}}_{j}$, together with the relevance vector $\boldsymbol{\omega}_{j}$ can be used as reject options \cite{chow}, determining whether or not an input pattern is in the receptive field of a winner node.

The overall operation of the map comprises three phases: 1) organization (\aref{alg:sssom2-algorithm}); 2) convergence; and 3) clustering or classification.

\begin{algorithm}[!ht]
\small
Initialize parameters \textit{lp}, $\beta$, \textit{age\_wins}, $e_{b}$, $e_{n}$, $s$, \textit{minwd}, $epoch_{max}$,  $N_{max}$;

Initialize the map with one node with $\boldsymbol{c}_{j}$ initialized at the first input pattern $\boldsymbol{x}_{0}$, $\boldsymbol{\omega}_{j}$ $\gets$ \textbf{1}, $\boldsymbol{\delta_j}$ $\gets$ \textbf{0}, $\hat{\boldsymbol{\delta}}_{j}$ $\gets$ \textbf{0}, $\text{t}_j$ $\gets$ 0, $\text{wins}_j$ $\gets$ 0 and $\text{class}_j$ $\gets$ \textit{noClass} or \textit{class}($\boldsymbol{x}_{0}$) if available;

Initialize the variable nwins $\gets$ 1;

\For{epoch $\gets$ 0 \textit{\textbf{to}} $epoch_{max}$}
{
	Choose a random input pattern $\boldsymbol{x}$;

	Compute the activation of all nodes (\eref{eq:larfdssom-activation});

	Find the winner $s_1$ with the highest activation (\eref{eq:larfdssom-winner});

	\eIf{\text{$\boldsymbol{x}$ has a label}}
    {
    	Run the SupervisedMode($\boldsymbol{x}$, $s_1$) (\aref{alg:sssom2-supervised});

    } {
    	Run the UnsupervisedMode($\boldsymbol{x}$, $s_1$) (\aref{alg:sssom2-unsupervised});
    }

    \If{\textit{nwins} = \textit{age\_wins}}
    {
    	Remove nodes with $\text{wins}_{j}$ $<$ \textit{lp} $\times$ \textit{age\_wins};

			Update the connections of the remaining nodes; 

      Reset the number of wins of the remaining nodes:

      $\text{wins}_{j}$ $\gets$ 0;

      \textit{nwins} $\gets$ 0;
    }

    \textit{nwins} $\gets$ \textit{nwins} + 1;
}

Run the Convergence Phase;
\caption{\gls{propm}}
\label{alg:sssom2-algorithm}
\end{algorithm}

In the organization phase, the network is initialized, and the nodes start to compete to form clusters of randomly chosen input patterns. The first node of the map is created at the same position of the first input pattern. As in \gls{sssom}, there are two distinct ways to define the winner of a competition, to decide when a new node must be inserted and when the nodes need to be updated. However, in \gls{propm}, before a node is updated, it is necessary to decide if it will affect the whole node structure or just the weighted averages and the relevance vectors. If the input pattern class label is provided, it will be done in the supervised mode (\sref{subsec:sup-mode}), otherwise, in the unsupervised mode (\sref{subsec:unsup-mode}).


The neighborhood of \gls{propm} is defined as the same as in \gls{sssom}. Nonetheless, it defines the nodes that will be adjusted together with the winner, thus outlining the cooperation step. The competition and cooperation steps are repeated for a limited number of epochs, and during this process, some nodes are removed periodically, conforming to a defined removal rule that a node must win at least for a minimum number of patterns to continue in the map, as in \gls{sssom}.

The convergence phase starts right after the organization process. In this phase, the nodes are also updated and removed when required, like in the way conducted in the first phase but with a slight difference: there is no insertion of new nodes. Finally, when the convergence phase finishes, the map clusters and classifies input patterns. At this stage, as in the \gls{sssom}, there are three possible scenarios: 1) all of the nodes have a defined representing class; 2) a mixed scenario, with some nodes labeled and other not; and 3) none of the nodes labeled. The first scenario will allow both classification and clustering tasks to be executed straightforwardly. The second will add one more step to the process because if the most activated node does not have a defined class, the algorithm continues trying to find a next highly activated node with a defined class. The last scenario only provides the ability to cluster.

\subsection{Structure of the Nodes}
\label{subsec:nodes-structure}

In \gls{propm}, each node \textit{j} in the map represents a cluster and is associated with four \textit{m}-dimensional vectors, where \textit{m} is the number of input dimensions: The first three vectors, $\boldsymbol{c}_j$, $\boldsymbol{\omega}_j$, and $\boldsymbol{\delta}_{j}$, are the same as defined in \cite{sssom}, , where $\textbf{c}_j = \{c_{ji}, i = 1, \cdot\cdot\cdot, m\}$ is the center vector; $\boldsymbol{\omega}_j = \{\omega_{ji}, i = 1, \cdot\cdot\cdot, m\}$ is the relevance vector; $\boldsymbol{\delta}_{j} = \{\delta_{ji}, i = 1, \cdot\cdot\cdot, m\}$ is the distance vector that stores moving averages of the observed distance between the input patterns \textbf{x} and the center vector $| \textbf{x} - \textbf{c}_j(n)|$ for each dimension. Note, however, that $\boldsymbol{\delta}$ in \gls{sssom} and \gls{propm} can be seen as the biased first moment estimate. Because of that, \gls{propm} introduces a fourth vector, $\hat{\boldsymbol\delta}_{j} = \{\hat{\delta}_{ji}, i = 1, \cdot\cdot\cdot, m\}$, which is the bias-corrected first moment estimate that the algorithm computes to counteract the bias towards zero of $\boldsymbol\delta_j$, specifically at the initial steps. The $\hat{\boldsymbol{\delta}}_{j}$ vector is used to compute the relevance vector $\boldsymbol{\omega}_j$, and both of them are used to approximate the variance of each node, taking into account how significant each dimension is. Such variance is used to define local reject options during the learning process every time that a new input pattern is presented to the map.

\subsection{Competition}
\label{subsec:competition}
\gls{propm} tries to choose the winner of a competition as the most activated node given an input pattern $\boldsymbol{x}$, except in certain cases that will be discussed in \sref{subsec:sup-mode}, when the label is available. In \gls{propm}, likewise in \gls{sssom}, the most activated node \textit{s}($\boldsymbol{x}$) is defined as per \eref{eq:larfdssom-winner}:
\begin{equation}
\small
\label{eq:larfdssom-winner}
s_1(\boldsymbol{x}) = \text{arg} \max_j [ac(D_{\omega}(\boldsymbol{x}, \boldsymbol{c}_j), \boldsymbol{\omega}_j)].
\end{equation}

where $\boldsymbol{\omega_{j}}$ is the relevance vector of the node \textit{j} and $ac(D_{\omega}(\boldsymbol{x}, \boldsymbol{c}_j)$ is the activation function.

As in \gls{sssom}, the activation function in \gls{propm} is calculated according to a radial basis function with the receptive field adjusted as a function of its relevance vector $\boldsymbol{\omega}_{j}$, as shown in \eref{eq:larfdssom-activation}:
\begin{equation}
\small
\label{eq:larfdssom-activation}
ac(D_{\omega}(\boldsymbol{x}, \boldsymbol{c}_j), \boldsymbol{\omega_j}) =  \frac{\sum\limits_{i=1}^m \omega_{ji}}{\sum\limits_{i=1}^m \omega_{ji} + D_{\omega}(\boldsymbol{x}, \boldsymbol{c}_j) + \epsilon},
\end{equation}

where $\epsilon$ is a small float number added to avoid division by zero and $D_{\omega}(\boldsymbol{x}, \boldsymbol{c}_j)$ is the weighted distance function:
\begin{equation}
\small
\label{eq:larfdssom-distance}
D_{\omega}(\boldsymbol{x}, \boldsymbol{c}_j) =  \sqrt{\sum_{i = 1}^{m} \omega_{ji} {(x_{i} - w_{ji})}^{2}}.
\end{equation}

\subsection{Estimating bias-corrected moving averages}
\label{subsec:estimate}
In \gls{propm}, the procedure that updates the distance vectors, as well as the relevance vectors, is shown in the \aref{alg:sssom2-update-relevances}.
\begin{algorithm}[!ht]
\small

\SetKwInOut{Input}{Input}
\Input{Input pattern $\boldsymbol{x}$, Node \textit{s}}

\SetKwFunction{FUpdate}{UpdateRelevances}
\SetKwProg{Function}{Function}{:}{}

\Function{\FUpdate{$\boldsymbol{x}$, \textit{s}}}
{
    Set $\text{t}_{s}$ $\gets$ $\text{t}_{s}$ + 1;

	Update the distance vector $\boldsymbol{\delta}_{s}$ (\eref{eq:sssom2-mv-avg});

	Update the corrected distance vector $\hat{\boldsymbol{\delta}_{s}}$ (\eref{eq:sssom2-mv-avg-corr});

	Update the relevance vector $\boldsymbol{\omega}_{s}$ (\eref{eq:sssom2-relevance});
}

\caption{Update Relevances}
\label{alg:sssom2-update-relevances}
\end{algorithm}

The distance vectors are initialized as a vector of zeros and updated through a moving average of the observed distance between the input pattern and the current center vector $\boldsymbol{c}_j$, as per \eref{eq:sssom2-mv-avg}:
\begin{equation}
\small
\label{eq:sssom2-mv-avg}
\boldsymbol{\delta}_j(n + 1) = \beta \boldsymbol{\delta}_j(n) + (1 - \beta)(|\boldsymbol{x} - \boldsymbol{c}_j(n)|),
\end{equation}

where $\beta \in \left]0, 1\right]$ is the parameter that controls the rate of change of the moving average (i.e., the exponential decay rate), and $| \boldsymbol{x} - \boldsymbol{c}_j(n) |$ denotes the absolute value applied to the elements of the vectors.

In order to correct the bias towards zero of $\boldsymbol\delta_j$ at the initial timesteps, caused by initializing the moving averages with zeros, as in the Adam algorithm \cite{adam}, \gls{propm} divides it by the term $\left( 1 - \beta^{t_j} \right)$, where $\textit{t}_j$ indicates the current timestep of each node \textit{j}. In sum, the bias-corrected moving averages vectors are updated at every node timestep according to the \eref{eq:sssom2-mv-avg-corr}:
\begin{equation}
\small
\label{eq:sssom2-mv-avg-corr}
\hat{\boldsymbol{\delta}}_j(n + 1) = \frac{\boldsymbol{\delta}_{j}(n)}{1 - \beta^{t_j}}
\end{equation}

To obtain accurate information about the relevance of each dimension for a given node, an update of the relevance vectors must follow every moving averages update. It is calculated by an inverse logistic function of the bias-corrected estimated distances $\hat{\boldsymbol{\delta}}_{ji}$, as follows in  \eref{eq:sssom2-relevance}.

\begin{equation}
\small
\label{eq:sssom2-relevance}
\omega_{ji} = \begin{cases}
  \frac{1}{1 \text{ + exp} \Big( \frac{\hat{\delta}_{ji\text{mean}} - \hat{\delta}_{ji}}{s(\hat{\delta}_{ji\text{max}} - \hat{\delta}_{ji\text{min}})} \Big)} & \text{if } \hat{\delta}_{ji\text{min}} \neq \hat{\delta}_{ji\text{max}} \\
  1 & \text{otherwise},
\end{cases}
\end{equation}

where $\hat{\delta}_{ji\text{max}}, \hat{\delta}_{ji\text{min}}$ and $\hat{\delta}_{ji\text{mean}}$ are respectively the maximum, the minimum and, the mean of the components of the bias-corrected moving average vector $\hat{\boldsymbol{\delta}}_{j}$, and the the parameter $s > 0$ controls the slope of the logistic function \cite{bassani2015larfdssom}. This function is pretty similar to the one used in \gls{sssom}, however, instead of using $\boldsymbol{\delta}_{j}$, the \gls{propm} replaces it by $\hat{\boldsymbol{\delta}}_{j}$ in order to get a more accurate and unbiased moving average value.

\subsection{Local Thresholds}
\label{subsec:local-threshold}

The distance vectors $\hat{\boldsymbol{\delta}}$ represent the corrected moving average of the observed distances between the input patterns $\boldsymbol{x}$ and center vectors $\boldsymbol{c}$ for each node \textit{j} in the map. As a result, they can be considered as the variances of the nodes, as stated before.

In addition, the $\boldsymbol{\omega}$ vectors express how each of the dimensions is important for each node, which indicates the subspaces of the input dimensions of a given dataset. This information corroborates with the definition of a local threshold, together with the estimated variance in the form of the $\hat{\boldsymbol{\delta}}$ vectors.

Combining them, it is possible to define a local region around each node center $\boldsymbol{c}_j$ to act like a reject option for some input patterns. If only the variances were used, some unimportant dimensions with a low variance could misguide the process when a similar input pattern $\boldsymbol{x}$ is outside the acceptance region of a node \textit{j}, but only in dimensions that are not relevant to it. Therefore, a flexible variance is defined to act as a local threshold and rejection option to mitigate such problems:
\begin{equation}
\small
\label{eq:sssom2-estimated-variance}
\text{Var}(\hat{\boldsymbol{\delta}}_{j}, \boldsymbol{\omega}_j) = \frac{\hat{\boldsymbol{\delta}}_{j}}{\boldsymbol{\omega}_j}
\end{equation}

When a dimension has a high relevance to the node, it will not impact its variance value. However, when a dimension has a small relevance, \gls{propm} will relax the constraints to allow a better definition of subspaces. Therefore, the general acceptance rule is defined by \eref{eq:sssom2-representativeness}, where the idea is to approximate to an optimal rule.
\begin{equation}
\small
\label{eq:sssom2-representativeness}
A(\boldsymbol{x}, \boldsymbol{c}_{j}, \boldsymbol{v}_{j}) = \begin{cases}
True,   & \boldsymbol{x}_i \in \left] \boldsymbol{c}_{ji} \pm \boldsymbol{v}_{ji} \right[,\\
        & \forall \textbf{ c}_{ji} \in \boldsymbol{c}_j \text{, }\boldsymbol{x}_{i} \in \boldsymbol{x} \text{, and } \boldsymbol{v}_{ji} \in \boldsymbol{v}_{j}\\

False, & \text{otherwise,}
 \end{cases}
\end{equation}

where $\boldsymbol{x}$, $\boldsymbol{c}_{j}$ are respectively the input pattern, and the center vector, and $\boldsymbol{v}_{j}$ = Var$(\hat{\boldsymbol{\delta}}_{j}, \boldsymbol{\omega}_j)$ is the relaxed variance vector as per \eref{eq:sssom2-estimated-variance}.

\subsection{Node Update}
\label{subsec:node-update}
As in \gls{larfdssom}, \gls{propm} updates the winner node and its neighbors using two distinct learning rates, $e_b$, and $e_n$, respectively. The \aref{alg:sssom2-update-node} shows how the whole update occurs.
\begin{algorithm}[!ht]
\small
\SetKwInOut{Input}{Input}
\Input{Input pattern $\boldsymbol{x}$, Node \textit{s}, Learning Rate \textit{lr}}

\SetKwFunction{FUpdate}{UpdateNode}
\SetKwProg{Function}{Function}{:}{}

\Function{\FUpdate{\textit{s}, \textit{lr}}}
{
	UpdateRelevances($\boldsymbol{x}$, s) (\aref{alg:sssom2-update-relevances});

	Update the weight vector $\boldsymbol{c}_{s}$ (\eref{eq:larfdssom-update-weights});
}

\caption{Update Node}
\label{alg:sssom2-update-node}
\end{algorithm}

The relevances and weighted moving averages are updated as shown in \sref{subsec:estimate}, and the centroid vector $\boldsymbol{c}_j$, given a learning rate \textit{lr}, is updated as follows:
\begin{equation}
\small
\label{eq:larfdssom-update-weights}
\boldsymbol{c}_j(n + 1) = \boldsymbol{c}_j(n) + e(\boldsymbol{x} - \boldsymbol{c}_j(n)),
\end{equation}

\subsection{Unsupervised Mode}
\label{subsec:unsup-mode}
Given an unlabeled input pattern, the most activated node is considered as the winner, disregarding its class labels. In this sense, \gls{propm} verifies if the condition expressed by \eref{eq:sssom2-representativeness} is satisfied. If so, the winner and its neighbors are updated towards the input pattern. Otherwise, a new node is inserted into the map at the input pattern position. However, since $s_1$ is the original winner, it will improve its knowledge about the region where it is located by updating its moving averages and relevances, but not its center. This mechanism provides the nodes the ability to learn about the region they are inserted in. An additional case is handled when the map has reached the maximum number of nodes. In this case, aiming at not losing the information that the input pattern can provide, as in previous models, and improving sample efficiency, \gls{propm} updates the moving average and the relevance vectors of the winning node. \aref{alg:sssom2-unsupervised} illustrates this procedure.
\begin{algorithm}[!ht]
\small
\SetKwInOut{Input}{Input}
\Input{Input pattern $\boldsymbol{x}$ and the first winner $s_1$;}
\If(\Comment{See \eref{eq:sssom2-representativeness}}){A($\boldsymbol{x}$, $\boldsymbol{c}_{s_1}$, Var($\hat{\boldsymbol{\delta}}_{s_1}$, $\boldsymbol{\omega}_{s_1}$)) is True and \textit{N} $<$ $\textit{N}_{max}$\\}
{
    Update the winner node and its neighbors: UpdateNode($s_1$, $e_b$), UpdateNode(\textit{neighbors}($s_1$), $e_n$) (\aref{alg:sssom2-update-node});

    Set $\text{wins}_{s_1}$ $\gets$ $\text{wins}_{s_1}$ + 1;
}
\ElseIf(\Comment{See \eref{eq:sssom2-representativeness}}){A($\boldsymbol{x}$, $\boldsymbol{c}_{s_1}$, Var($\hat{\boldsymbol{\delta}}_{s_1}$, $\boldsymbol{\omega}_{s_1}$)) is False}
{
    Create a new node \textit{j} and set: $\boldsymbol{c}_{j}$ $\gets$ $\boldsymbol{x}$, $\boldsymbol{\omega}_j$ $\gets$ \textbf{1}, $\boldsymbol{\delta}_{j}$ $\gets$ \textbf{0}, $\hat{\boldsymbol{\delta}}_{j}$ $\gets$ \textbf{0}, $t_j$ $\gets$ 0, $\text{class}_{j}$ $\gets$ \textit{noClass} and $\text{wins}_j$ $\gets$ $lp \times nwins$;

    Connect j to its neighbors; 

    Update the relevances vector of $s_1$: UpdateRelevances($\boldsymbol{x}$, $s_1$) (\aref{alg:sssom2-update-relevances});
}
\Else
{
    Update the relevances vector of $s_1$: UpdateRelevances($\boldsymbol{x}$, $s_1$) (\aref{alg:sssom2-update-relevances});
}
\caption{Unsupervised Mode}
\label{alg:sssom2-unsupervised}
\end{algorithm}

\subsection{Supervised Mode}
\label{subsec:sup-mode}
\aref{alg:sssom2-supervised} shows how this supervised procedure is conducted. In this procedure, unlike the unsupervised mode, the labels are taken into account when looking for a winner. If the most activated node $s_1$ has the same class of the input pattern or a not defined class, a very similar approach to the unsupervised mode is applied. The difference is directly related to the fact that is necessary to set $s_1$ class as the same class of the given input pattern $\boldsymbol{x}$, as well as to update its connections. Otherwise, the \gls{propm} tries to find a new winner with the same class of the input pattern $\boldsymbol{x}$ or a not yet defined class.

If some new node takes the place of $s_1$ as a new winner $s_2$, the acceptance criteria expressed by the \eref{eq:sssom2-representativeness} is verified. If so, and the map is not full, the new winner and its neighbors are updated. Otherwise, only the moving averages and relevance vector of $s_2$ are updated in order to give the same chance received by $s_1$ to improve its knowledge about the surrounding area.

If there are no new nodes to replace $s_1$ as a new winner, and the map is not full, the $s_1$ node is duplicated, preserving the moving averages vectors, the centroid vector as well as the relevance vector. However, the class of this new duplicated node is set to the same as the input pattern. The other parameters are set as usual. If none of the above conditions are fulfilled, the \gls{propm} solely updates the moving averages and relevance vector of the first defined winner $s_1$.
\begin{algorithm}[!ht]
\small
\SetKwInOut{Input}{Input}
\Input{Input pattern $\boldsymbol{x}$ and the first winner $s_1$;}

\eIf{$\text{class}_{s_1}$ = \textit{class}($\boldsymbol{x}$) \textbf{or} $\text{class}_{s_1}$ = noClass}
{
	\If(\Comment{See \eref{eq:sssom2-representativeness}}){A($\boldsymbol{x}$, $\boldsymbol{c}_{s_1}$, Var($\hat{\boldsymbol{\delta}}_{s_1}$, $\boldsymbol{\omega}_{s_1}$)) is False and \textit{N} $<$ $\textit{N}_{max}$\\}
	{

    	Create new node \textit{j} and set: $\boldsymbol{c}_{j}$ $\gets$ $\boldsymbol{x}$, $\boldsymbol{\omega}_j$ $\gets$ \textbf{1}, $\boldsymbol{\delta}_{j}$ $\gets$ \textbf{0}, $\hat{\boldsymbol{\delta}}_{j}$ $\gets$ \textbf{0}, $\text{t}_j$ $\gets$ 0, $\text{class}_{j}$ $\gets$ \textit{class}($\boldsymbol{x}$) and $\text{wins}_j$ $\gets$ $lp \times nwins$;

			Connect j to its neighbors; 

	    Update the relevances vector of $s_1$: UpdateRelevances($\boldsymbol{x}$, $s_1$) (\aref{alg:sssom2-update-relevances});
	}
    \ElseIf(\Comment{See \eref{eq:sssom2-representativeness}}){A($\boldsymbol{x}$, $\boldsymbol{c}_{s_1}$, Var($\hat{\boldsymbol{\delta}}_{s_1}$, $\boldsymbol{\omega}_{s_1}$)) is True}
    {
    	Update the winner node and its neighbors: UpdateNode($s_1$, $e_b$), UpdateNode(\textit{neighbors}($s_1$), $e_n$) (\aref{alg:sssom2-update-node});

    	Set $\text{class}_{s_1}$ $\gets$ \textit{class}($\boldsymbol{x}$);

			Update $s_1$ connections; 

    	Set $\text{wins}_{s_1}$ $\gets$ $\text{wins}_{s_1}$ + 1;
    }
    \Else
    {
        Update the relevances vector of $s_1$: UpdateRelevances($\boldsymbol{x}$, $s_1$)
    }
}{
	Try to find a new winner $s_2$ as the next node with highest activation and \textit{noClass} or the same class of $\boldsymbol{x}$;

    \If{$s_2$ exists}
    {
        \If(\Comment{See \eref{eq:sssom2-representativeness}}){A($\boldsymbol{x}$, $\boldsymbol{c}_{s_2}$, Var($\hat{\boldsymbol{\delta}}_{s_2}$, $\boldsymbol{\omega}_{s_2}$)) is True and \textit{N} $<$ $\textit{N}_{max}$\\}
        {
    	    Update the new winner node and its neighbors: UpdateNode($s_2$, $e_b$) and  UpdateNode(\textit{neighbors}($s_2$), $e_n$) (\aref{alg:sssom2-update-node});
    	}
        \Else
        {
            Update the relevances vector of $s_2$: UpdateRelevances($\boldsymbol{x}$, $s_2$)
        }
        Set $\text{wins}_{s_2}$ $\gets$ $\text{wins}_{s_2}$ + 1;
    }
    \ElseIf{\textit{N} $<$ $\textit{N}_{max}$}
    {
    	Create new node \textit{j} by duplicating $s_1$ and set: $\boldsymbol{c}_{j}$ $\gets$ $\boldsymbol{c}_{s_1}$, $\boldsymbol{\omega}_j$ $\gets$ $\boldsymbol{\omega}_{s_1}$, $\boldsymbol{\delta}_{j}$ $\gets$ $\boldsymbol{\delta}_{s_1}$, $\hat{\boldsymbol{\delta}}_{j}$ $\gets$ $\hat{\boldsymbol{\delta}}_{s_1}$, $\text{t}_j$ $\gets$ 0, $\text{class}_{j}$ $\gets$ \textit{class}($\boldsymbol{x}$) and $\text{wins}_j$ $\gets$ $lp \times nwins$;

			Connect $j$ to its neighbors; 
    }
    \Else
    {
        Update the relevances vector of $s_1$: UpdateRelevances($\boldsymbol{x}$, $s_1$)
    }

}

\caption{Supervised Mode}
\label{alg:sssom2-supervised}
\end{algorithm}

\subsection{Node Removal}
\label{subsec:removal}
In \gls{propm}, as in \gls{sssom} and \gls{larfdssom}, each node \textit{j} stores a variable $wins_j$ that accounts for the number of nodes victories since the last reset. Whenever \textit{nwins} reaches the \textit{age\_wins} value, a reset occurs. It implies to the moment when nodes that did not win at least the minimum percentage of the competition $lp \times age\_wins$ are removed from the map. After a reset, the number of victories of the remaining nodes is reset to zero. Finally, to avoid the removal of recently created nodes, when a new node is inserted, its number of wins is set to $lp \times nwins$, where \textit{nwins} indicates the number of competitions that have occurred since the last reset.

\subsection{Parameters Summary}
\label{subsec:params}
\gls{propm} removes two parameters from its predecessor, \gls{sssom}. First, the parameter $a_t$, that has a great impact on the results as shown by \cite{bassani2015larfdssom}. It was replaced by the adaptive local threshold technique introduced by \gls{propm} (\sref{subsec:local-threshold}) that can define and learn the space region that a node can represent during the training. Second, the parameter $e_w$ was also removed due to its irrelevance in the learning process after removing $a_t$, i.e., \gls{propm} has nine parameters to be set up. More precisely, a sensitivity analysis revealed that there is no parameter with a high impact on the results anymore. This method seeks to establish a good level of self-adjustment, in a way that we can keep the parameters values fixed inside predefined ranges.

\section{Experiments}
\label{sec:experiments}

The experiments are divided into three parts. First, in order to evaluate the classification rate of \gls{propm}, we replicated the experiments conducted in \cite{sssom}, adding the proposed method to the comparison. Second, we compare the performance of \gls{doc}, \gls{proclus}, \gls{larfdssom}/\gls{sssom} and \gls{propm}. Remark that the first two methods are originally from the data mining area. This choice of comparison methods was defined taking into consideration the analysis provided by \cite{bassani2015larfdssom}, where \gls{larfdssom} presented the best results overall, and \gls{doc} and \gls{proclus} appeared as the two best options on average concerning subspace approaches in a distinction of data mining applications. Also, we refer the \gls{larfdssom} and \gls{sssom} together due to their equivalence for clustering tasks solely. Third, we performed a sensitivity analysis to show that with \gls{propm} the same range of parameters work well for a variety of datasets and that range does not exist for the previous methods (\gls{larfdssom} and \gls{sssom}).

For all experiments, the seven real-world datasets provided by the OpenSubspace framework \cite{muller2009evaluating} were used, rescaling them to the [0, 1] interval. \sref{subsec:exp-setup} presents the experimental setup that was used. Next in the \sref{subsec:exp-results}, we present the results and analysis that are necessary to clarify the conclusions taken.

\subsection{Experimental Setup}
\label{subsec:exp-setup}
For the first set of experiments, on every dataset, we used 3-times 3-fold cross-validation to measure the classification rate, as in \cite{sssom}. Still as in \cite{sssom}, for studying the effects of the different the percentage of labeled data, the semi-supervised methods were trained with 1\%, 5\%, 10\%, 25\%, 50\%, 75\% and 100\% of labels. In the second group of experiments, we have chosen the \gls{ce} metric, as in \cite{bassani2015larfdssom} to evaluate the clustering assignments. For that, we considered all dimensions as relevant in the target clusterings used to calculate the \gls{ce}. Also as in \cite{bassani2015larfdssom}, we considered a problem of projected clustering, where each sample is assigned to a single cluster. Finally, we perform a sensitivity analysis with \gls{larfdssom} and \gls{propm} to elucidate the gains obtained regarding the importance that a parameter has in the outcome result. It will establish \gls{propm} as a more robust and self-controlled model.

For all the experiments, we sampled the parameter ranges according to a \gls{lhs} \cite{helton2005comparison}, that guarantees the full coverage of the range of each parameter. In this sense, we gathered 500 different parameter settings, i.e., the range of each parameter was divided into 500 intervals of equal probability, resulting in a random selection of a single value from each interval \cite{helton2005comparison}. The ranges used for \gls{propm} are defined in \tref{tab:sssom-params}, whereas the ranges of the other methods were the same as those used in \cite{bassani2015larfdssom,sssom}. We set the maximum number of nodes for \gls{propm} to be 200.

\begin{table}[ht!]
\centering
\begin{threeparttable}
\renewcommand{\arraystretch}{1.3}
\caption{Parameter Ranges for \gls{propm}}
\label{tab:sssom-params}
\centering
\begin{tabular}{lcc}
\hline
\bfseries  Parameters & \bfseries min & \bfseries max\\
\hline
Lowest cluster percentage (lp) & 0.001 & 0.002 \\
Relevance rate ($\beta$) & 0.90 & 0.95 \\
Max competitions ($age\_wins$) & $1 \times S^*$ & $200 \times S^*$ \\
Winner learning rate ($e_b$) & 0.001 & 0.2 \\
Neighbors learning rate ($e_n$) & $0.002 \times e_b$ & $ 1 \times e_b$ \\
Relevance smoothness ($\epsilon \beta$) & 0.01 & 0.1 \\
Connection threshold ($minwd$) & 0 & 0.5 \\
Number of epochs ($epochs$) & 1 & 100 \\
\hline
\end{tabular}
\begin{tablenotes}
\small\item * \textit{S} is the number of input patterns in the dataset.
\end{tablenotes}
\end{threeparttable}
\end{table}

\subsection{Experimental Results and Analysis}
\label{subsec:exp-results}

\begin{figure*}[ht!]
  \centering
  \subfigure[Breast]{\includegraphics[width=0.32\linewidth,scale=1]{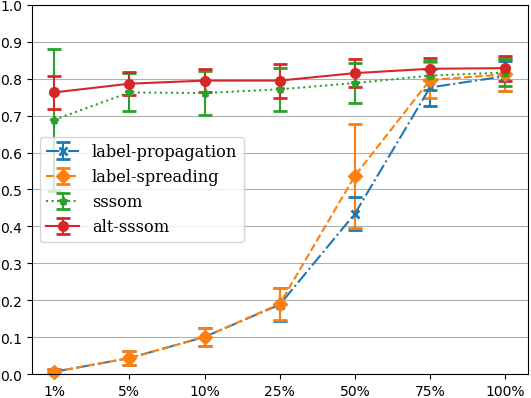}
  \label{fig:breast}}
  \subfigure[Diabetes]{\includegraphics[width=0.32\linewidth,scale=1]{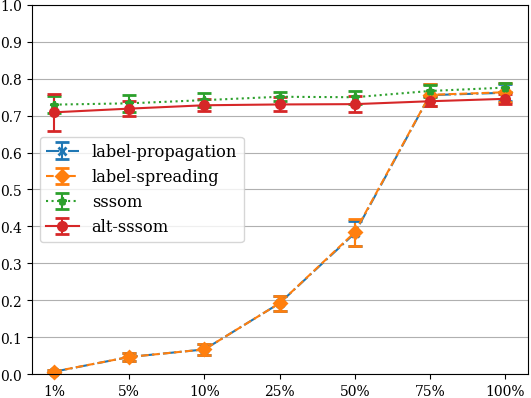}
  \label{fig:diabetes}}
  \subfigure[Glass]{\includegraphics[width=0.32\linewidth,scale=1]{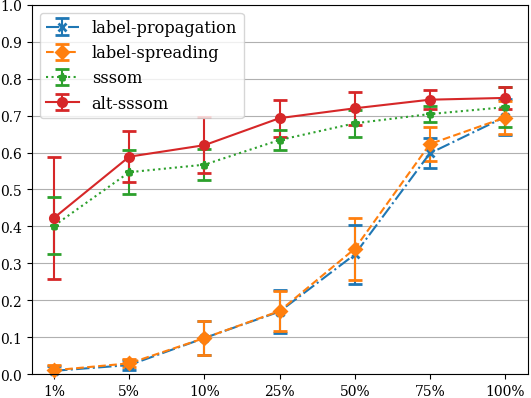}
  \label{fig:glass}}

  \centering
  \subfigure[Shape]{\includegraphics[width=0.32\linewidth,scale=1]{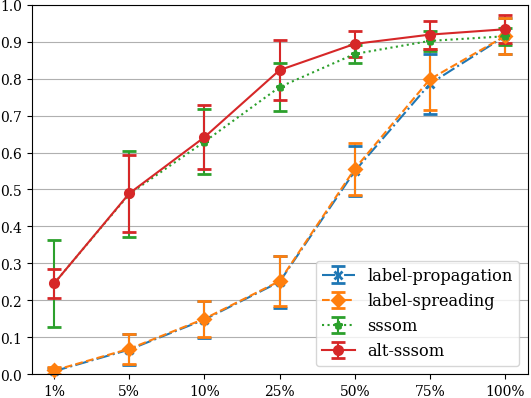}
  \label{fig:shape}}
  \subfigure[Pendigits]{\includegraphics[width=0.32\linewidth,scale=1]{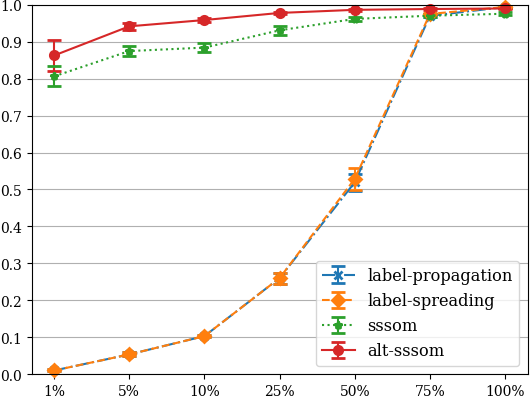}
  \label{fig:pendigits}} 
  \subfigure[Vowel]{\includegraphics[width=0.32\linewidth,scale=1]{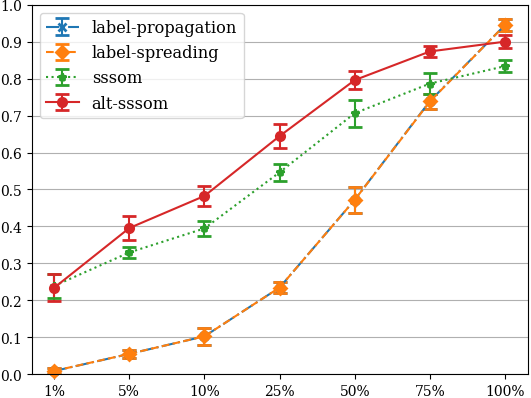}
  \label{fig:vowel}}


  \caption{Best mean accuracy and standard deviation as function of the percentage of supervision for each dataset}
  \label{fig:plots}
\end{figure*}

\fref{fig:plots} shows the results of \gls{propm} in comparison with \gls{sssom}, Label Propagation and Label Spreading in the real-world datasets. The results are shown as a function of the percentage of labels that were used. Overall, the \gls{propm} improved the performance of \gls{sssom}, except for the Diabetes dataset (\fref{fig:diabetes}) where the results obtained were slightly worse, but yet comparable. The flexibility provided by the estimation of the representing area allowed such results. Still, the standard deviation for all datasets in all supervision levels was also minimized, which indicates another positive aspect of the proposed method: it is more robust to variations on both datasets and parameters. With \gls{sssom}, the other semi-supervised methods surpassed its performance in the Pendigits and Vowel datasets, however, \gls{propm} achieved a consistent improvement by outperforming the results of both Label Propagation and Spreading in such cases. On all the other situations of this experiment, the \gls{propm} outperformed the comparing models.

Second, \tref{tab:results} shows the results of the \gls{ce} obtained with the methods. It shows that no method achieved the best result for all real-world datasets. Even though the \gls{propm} presented an overall result that is better than all the others, it achieved the same results of \gls{doc} and \gls{larfdssom} for the Breast dataset. For the Glass dataset, again, it showed the same result of \gls{larfdssom}. Also, \gls{propm} was not the best for Diabetes, which shows a similar behavior when compared with the results obtained in the first set of experiments that took into account the classification rate. On considering a general comparison, \gls{propm} present the best results on average. It is worth mentioning that the similarity between the results of \gls{propm} and \gls{larfdssom} can be attributed to the fact that there were no labeled noise samples in the datasets, to the unknown information about the irrelevant dimensions, and to the intrinsic characteristics inherited by \gls{propm} from the \gls{larfdssom}. Also, once \gls{doc} does not have a direct way to control the number of clusters, it displays some difficulty to find out the correct value.
Moreover, \gls{proclus} presented good results when the parameter controlling the number of clusters is defined close to the optimum. The good results obtained by \gls{larfdssom} is directly related to an excellent choice of the parameters $a_t$ and \textit{lp}, which significantly impact the results. \gls{propm} achieves a good result without the needing of a highly accurate definition of parameters, as well as it is not necessary to define an a priori number of clusters due to its time-varying feature.

\begin{table*}[ht]
\small
\centering
\renewcommand{\arraystretch}{1.3}
\caption{CE Results for Real-World Datasets. Best results for each dataset are shown in bold}
\label{tab:results}
\centering
\begin{tabular}{c||lllllll|cc}
\hline
\bfseries  CE & \bfseries Breast & \bfseries Diabetes & \bfseries Glass & \bfseries Liver & \bfseries Pendigits & \bfseries Shape & \bfseries Vowel & \bfseries Average & \bfseries STD\\
\hline\hline
DOC & \textbf{0.763} & 0.654 & 0.439 & 0.580 & 0.566 & 0.419 & 0.142 & 0.509 & 0.201 \\
PROCLUS & 0.702 & 0.647 & 0.528 & 0.565 & 0.615 & 0.706 & 0.253 & 0.574 & 0.156 \\
LARFDSSOM & \textbf{0.763}  & \textbf{0.727} & \textbf{0.575} & 0.580 & 0.737 & 0.719 & 0.317 & 0.631 & 0.158 \\
ALT-SSSOM & \textbf{0.763}  & 0.697 & \textbf{0.575} & \textbf{0.603} & \textbf{0.741} & \textbf{0.738} & \textbf{0.319} & \textbf{0.633} & 0.156 \\
\hline
\end{tabular}
\end{table*}

\begin{figure*}[ht!]
  \centering
  \subfigure[$lp$ - Breast]{\includegraphics[width=0.23\linewidth,scale=1,trim=0mm 0mm 0mm 4.2mm,clip]{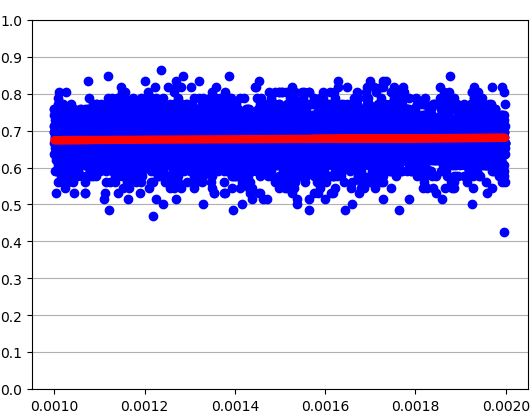}
  \label{fig:breast-lp}}
  \subfigure[$lp$ - Diabetes]{\includegraphics[width=0.23\linewidth,scale=1,trim=0mm 0mm 0mm 4.2mm,clip]{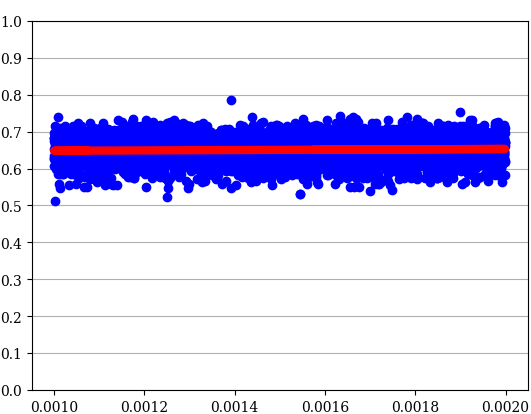}
  \label{fig:diabetes-lp}}
  \subfigure[$lp$ - Pendigits]{\includegraphics[width=0.23\linewidth,scale=1,trim=0mm 0mm 0mm 4.2mm,clip]{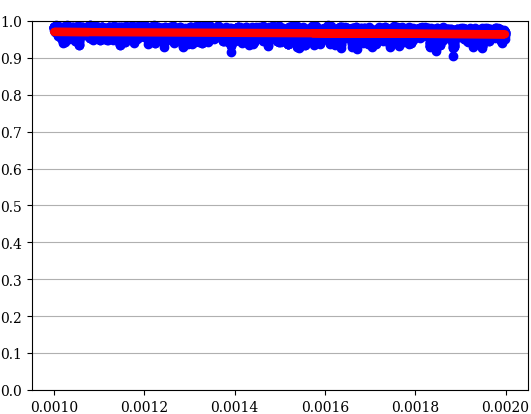}
  \label{fig:pendigits-lp}}
  \subfigure[$lp$ - Vowel]{\includegraphics[width=0.23\linewidth,scale=1,trim=0mm 0mm 0mm 4.2mm,clip]{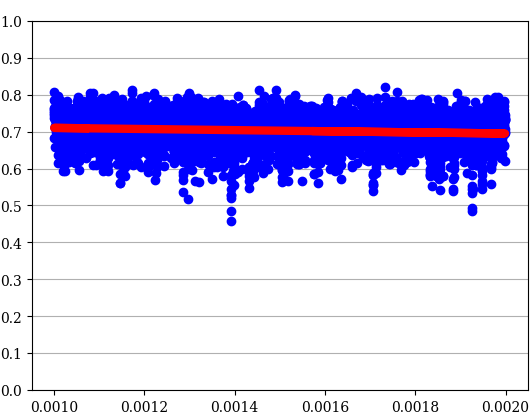}
  \label{fig:vowel-lp}}

  \caption{Scatter plots of the Accuracy obtained with \gls{propm} as a function of its parameter \textit{lp} for the datasets trained with 50\% of labels}
  \label{fig:plots-lp}
\end{figure*}

Third, \fref{fig:plots-lp} shows the scatter plot of the accuracy as a function of the parameter \textit{lp} for the datasets trained with 50\% of labels to illustrate a scenario where both forms of learning impact the outcome. Note that for all datasets, \textit{lp} did not show a significant impact on the results. The linear fit to the data, represented by the red line, highlight this by not exhibiting any trend. It is also worth mentioning that the plots in \fref{fig:plots-lp} are the combination of each parameter value for each cross-validation set. In previous versions, the most two critical parameters were $a_t$ and \textit{lp}. The parameter $a_t$ played a role of great importance due to its high impact on the results with just a small change on its values, i.e., $a_t$ impacted the results exponentially. Here in \gls{propm}, \textit{lp} is the most important parameter because it defines more clearly the behavior of the algorithm. Despite it, it does not impact the result.

\begin{figure*}[ht!]
  \centering
  \subfigure[$e_b$ - Breast]{\includegraphics[width=0.23\linewidth,scale=1,trim=0mm 0mm 0mm 4.2mm,clip]{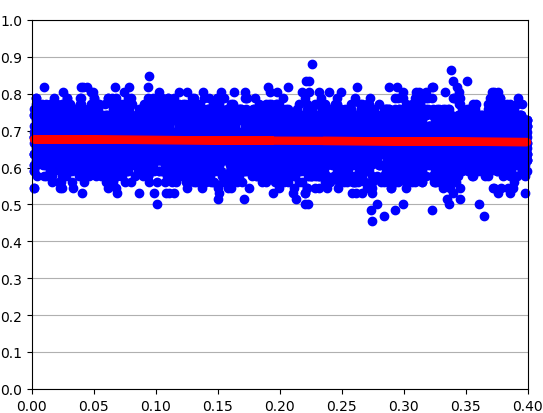}
  \label{fig:breast-eb04}}
  \subfigure[$e_b$ - Pendigits]{\includegraphics[width=0.23\linewidth,scale=1,trim=0mm 0mm 0mm 4.2mm,clip]{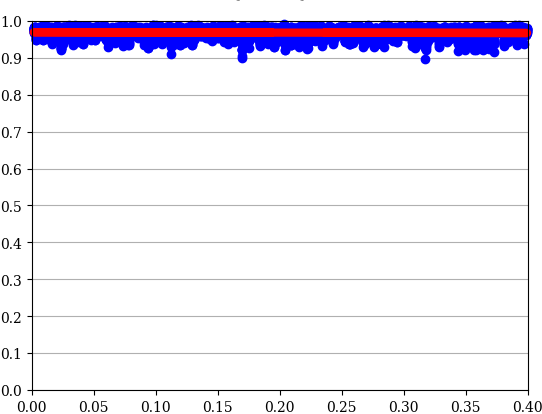}
  \label{fig:pendigits-eb04}}
  \subfigure[$e_b$ - Breast]{\includegraphics[width=0.23\linewidth,scale=1,trim=0mm 0mm 0mm 4.2mm,clip]{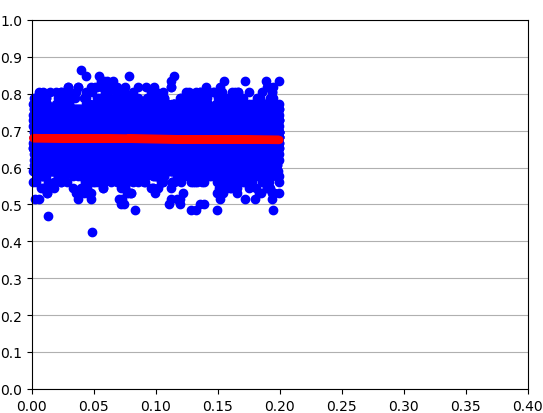}
  \label{fig:breast-eb02}}
  \subfigure[$e_b$ - Pendigits]{\includegraphics[width=0.23\linewidth,scale=1,trim=0mm 0mm 0mm 4.2mm,clip]{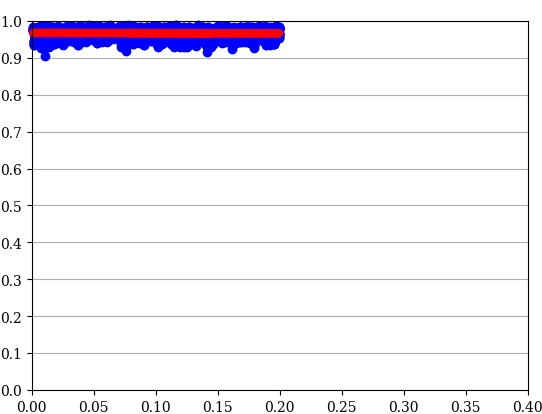}
  \label{fig:pendigits-eb02}}


  \caption{Scatter plots of the Accuracy obtained with \gls{propm} as a function of its parameter $e_b$ for the datasets trained with 50\% of labels to illustrate how the parameter ranges were defined}
  \label{fig:plots-eb}
\end{figure*}

\fref{fig:plots-eb} also shows a scatter plot of $e_b$ parameter. However, we pick it up the datasets Breast and Pendigits to illustrate the choice of parameters. We first started with a wide range from 0.001 to 0.4 (\fref{fig:breast-eb04} and \fref{fig:pendigits-eb04}). However, the linear fit was mostly horizontal, not indicating any trend, again. We then shrank the range to 0.001 to 0.2 and reran the experiments. The results were as expected, keeping as stable as shown by \fref{fig:breast-eb02} and \fref{fig:pendigits-eb02}. These experiments clarify how robust the model is to parameter changes, that is why we kept the same scale in the graphics. The two parameters taken for study in these figures were chosen due to its semantical importance, since other parameters presented similar behavior, with none of them acting a role as $a_t$ and \textit{lp} in the before-mentioned versions.

\section{Conclusion and Future Work}
\label{sec:conclusions}

This paper presented our second approach for semi-supervised self-organizing maps applied to cluster and classification tasks. The behavior of \gls{propm} shown to have led to improvements on its previous version, the \gls{sssom} not only in terms of classification rate but also in clustering aspects. It consolidates the position of a good choice in situations where only a small portion of labels are available. The clustering task also achieved a significant improvement with the proposed changes. The \gls{propm} was able to reduce in two the number of parameters, whereby improving the performance in both contexts of classification and clustering metrics.

Also, probably one of the most important contributions of this current paper is related to the parametric robustness showed by the third and final experiment. It is of great importance to reduce the dependency and the variability of models to some parameters, and \gls{propm} achieves this.

The use of a relaxed estimated variance and region allows the method to better explore the information available on the data, improving sample efficiency. Also, the modifications proposed in \gls{propm} provided the ability to not merely discard data in certain cases but keep digging into its characteristics in order to establish a better estimate of their statistics.

Finally, we have left for future work some promising approaches that are related to defining a better stop criterion, to the use of the unsupervised error to build a model with more than one layer, as well as a hierarchical approach to provide better exploitation of the data statistics.

\section*{ACKNOWLEDGMENTS}
The authors would like to thank CNPq (Conselho Nacional de Desenvolvimento Científico e Tecnológico), Brazil, for supporting this research study, and FACEPE (Fundação de Amparo à Ciência e Tecnologia do Estado de Pernambuco), Brazil, for financial support on the project \#APQ-0880-1.03/14.

\bibliographystyle{IEEEtran}
\bibliography{default_content/bibliography}

\begin{thebibliography}{10}
\providecommand{\url}[1]{#1}
\csname url@samestyle\endcsname
\providecommand{\newblock}{\relax}
\providecommand{\bibinfo}[2]{#2}
\providecommand{\BIBentrySTDinterwordspacing}{\spaceskip=0pt\relax}
\providecommand{\BIBentryALTinterwordstretchfactor}{4}
\providecommand{\BIBentryALTinterwordspacing}{\spaceskip=\fontdimen2\font plus
\BIBentryALTinterwordstretchfactor\fontdimen3\font minus
  \fontdimen4\font\relax}
\providecommand{\BIBforeignlanguage}[2]{{%
\expandafter\ifx\csname l@#1\endcsname\relax
\typeout{** WARNING: IEEEtran.bst: No hyphenation pattern has been}%
\typeout{** loaded for the language `#1'. Using the pattern for}%
\typeout{** the default language instead.}%
\else
\language=\csname l@#1\endcsname
\fi
#2}}
\providecommand{\BIBdecl}{\relax}
\BIBdecl

\bibitem{lecun2015deep}
Y.~LeCun, Y.~Bengio, and G.~Hinton, ``Deep learning,'' \emph{Nature}, vol. 521,
  no. 7553, p. 436, 2015.

\bibitem{sssom}
P.~H. Braga and H.~F. Bassani, ``A semi-supervised self-organizing map for
  clustering and classification,'' in \emph{2018 International Joint Conference
  on Neural Networks (IJCNN)}.\hskip 1em plus 0.5em minus 0.4em\relax IEEE,
  2018, pp. 1--8.

\bibitem{chapelle2009semi}
O.~Chapelle, B.~Scholkopf, and A.~Zien, ``Semi-supervised learning,''
  \emph{IEEE Transactions on Neural Networks}, vol.~20, no.~3, pp. 542--542,
  2009.

\bibitem{schwenker2014pattern}
F.~Schwenker and E.~Trentin, ``Pattern classification and clustering: A review
  of partially supervised learning approaches,'' \emph{Pattern Recognition
  Letters}, vol.~37, pp. 4--14, 2014.

\bibitem{kohonen1990}
T.~Kohonen, ``The self-organizing map,'' \emph{Proceedings of the IEEE},
  vol.~78, no.~9, pp. 1464--1480, 1990.

\bibitem{bassani2015larfdssom}
H.~F. Bassani and A.~F. Araujo, ``Dimension selective self-organizing maps with
  time-varying structure for subspace and projected clustering,'' \emph{IEEE
  transactions on neural networks and learning systems}, vol.~26, no.~3, pp.
  458--471, 2015.

\bibitem{zhu2002-label-propagation}
Z.~Xiaojin and G.~Zoubin, ``Learning from labeled and unlabeled data with label
  propagation,'' \emph{Tech. Rep., Technical Report CMU-CALD-02--107, Carnegie
  Mellon University}, 2002.

\bibitem{csom}
H.~Dozono, G.~Niina, and S.~Araki, ``Convolutional self organizing map,'' in
  \emph{CSCI}.\hskip 1em plus 0.5em minus 0.4em\relax IEEE, 2016, pp. 767--771.

\bibitem{chen2018semi}
L.~Chen, S.~Yu, and M.~Yang, ``Semi-supervised convolutional neural networks
  with label propagation for image classification,'' in \emph{2018 24th
  International Conference on Pattern Recognition (ICPR)}.\hskip 1em plus 0.5em
  minus 0.4em\relax IEEE, 2018, pp. 1319--1324.

\bibitem{rasmus2015semi}
A.~Rasmus, M.~Berglund, M.~Honkala, H.~Valpola, and T.~Raiko, ``Semi-supervised
  learning with ladder networks,'' in \emph{Advances in Neural Information
  Processing Systems}, 2015, pp. 3546--3554.

\bibitem{fischer2016optimal}
L.~Fischer, B.~Hammer, and H.~Wersing, ``Optimal local rejection for
  classifiers,'' \emph{Neurocomputing}, vol. 214, pp. 445--457, 2016.

\bibitem{adam}
D.~P. Kingma and J.~Ba, ``Adam: A method for stochastic optimization,''
  \emph{arXiv preprint arXiv:1412.6980}, 2014.

\bibitem{chow}
C.~Chow, ``On optimum recognition error and reject tradeoff,'' \emph{IEEE
  Transactions on information theory}, vol.~16, no.~1, pp. 41--46, 1970.

\bibitem{fischer2014rejection}
L.~Fischer, B.~Hammer, and H.~Wersing, ``Rejection strategies for learning
  vector quantization.'' in \emph{ESANN}, 2014.

\bibitem{koppen2000curse}
M.~K{\"o}ppen, ``The curse of dimensionality,'' in \emph{5th Online World
  Conference on Soft Computing in Industrial Applications}, 2000, pp. 4--8.

\bibitem{kriegel2009clustering}
H.-P. Kriegel, P.~Kr{\"o}ger, and A.~Zimek, ``Clustering high-dimensional data:
  A survey on subspace clustering, pattern-based clustering, and correlation
  clustering,'' \emph{ACM Transactions on Knowledge Discovery from Data},
  vol.~3, no.~1, p.~1, 2009.

\bibitem{vailaya2000reject}
A.~Vailaya and A.~Jain, ``Reject option for vq-based bayesian classification,''
  in \emph{ICPR}.\hskip 1em plus 0.5em minus 0.4em\relax IEEE, 2000, p. 2048.

\bibitem{adaptive-local-thresholding}
X.~Jiang and D.~Mojon, ``Adaptive local thresholding by verification-based
  multithreshold probing with application to vessel detection in retinal
  images,'' \emph{IEEE Transactions on Pattern Analysis and Machine
  Intelligence}, vol.~25, no.~1, pp. 131--137, 2003.

\bibitem{singh2012new}
T.~R. Singh, S.~Roy, O.~I. Singh, T.~Sinam, K.~Singh \emph{et~al.}, ``A new
  local adaptive thresholding technique in binarization,'' \emph{arXiv preprint
  arXiv:1201.5227}, 2012.

\bibitem{label-spreading}
D.~Zhou, O.~Bousquet, T.~N. Lal, J.~Weston, and B.~Sch{\"o}lkopf, ``Learning
  with local and global consistency,'' in \emph{Advances in neural information
  processing systems}, 2004, pp. 321--328.

\bibitem{deep-ssl}
Z.~Hailat, A.~Komarichev, and X.~Chen, ``Deep semi-supervised learning,'' in
  \emph{2018 24th International Conference on Pattern Recognition (ICPR)}, Aug
  2018, pp. 2154--2159.

\bibitem{dsom}
N.~Liu, J.~Wang, and Y.~Gong, ``Deep self-organizing map for visual
  classification,'' in \emph{Neural Networks (IJCNN), 2015 International Joint
  Conference on}.\hskip 1em plus 0.5em minus 0.4em\relax IEEE, 2015, pp. 1--6.

\bibitem{basu2002semi}
S.~Basu, A.~Banerjee, and R.~Mooney, ``Semi-supervised clustering by seeding,''
  in \emph{In Proceedings of 19th International Conference on Machine
  Learning}, 2002.

\bibitem{jain2010data}
A.~K. Jain, ``Data clustering: 50 years beyond k-means,'' \emph{Pattern
  recognition letters}, vol.~31, no.~8, pp. 651--666, 2010.

\bibitem{proclus}
C.~C. Aggarwal, J.~L. Wolf, P.~S. Yu, C.~Procopiuc, and J.~S. Park, ``Fast
  algorithms for projected clustering,'' in \emph{ACM SIGMoD Record}, vol.~28,
  no.~2.\hskip 1em plus 0.5em minus 0.4em\relax ACM, 1999, pp. 61--72.

\bibitem{doc}
C.~M. Procopiuc, M.~Jones, P.~K. Agarwal, and T.~Murali, ``A monte carlo
  algorithm for fast projective clustering,'' in \emph{Proceedings of the 2002
  ACM SIGMOD international conference on Management of data}.\hskip 1em plus
  0.5em minus 0.4em\relax ACM, 2002, pp. 418--427.

\bibitem{muller2009evaluating}
E.~M{\"u}ller, S.~G{\"u}nnemann, I.~Assent, and T.~Seidl, ``Evaluating
  clustering in subspace projections of high dimensional data,''
  \emph{Proceedings of the VLDB Endowment}, vol.~2, no.~1, pp. 1270--1281,
  2009.

\bibitem{helton2005comparison}
J.~C. Helton, F.~Davis, and J.~D. Johnson, ``A comparison of uncertainty and
  sensitivity analysis results obtained with random and latin hypercube
  sampling,'' \emph{Reliability Engineering \& System Safety}, vol.~89, no.~3,
  pp. 305--330, 2005.

\end{thebibliography}

\end{document}